**Chapter Author(s)**	Teena, Arora[*]

Federation University, Mt Helen, Australia, teenaarora@students.federation.edu.au

Venki, Balasubramanian

Federation University, Mt Helen, Australia, v.balasubramanian@federation.edu.au

Andrew, Stranieri

Federation University, Mt Helen, Australia, a.stanieri@federation.edu.au

Shenhan, Mai

Federation University, Mt Helen, Australia, shenhanmai@students.federation.edu.au

Rajkumar, Buyya

The University of Melbourne, Australia, rbuyya@unimelb.edu.au

Sardar, Islam

Victoria University, sardar.islam@vu.edu.au

**Chapter Title**	Classification of Methods to Reduce Clinical Alarm Signals for Remote Patient Monitoring: A Critical Review

**Abstract**

Remote Patient Monitoring (RPM) is an emerging technology paradigm that helps reduce clinicians' workload by automated monitoring and raising intelligent alarm signals. High sensitivity and intelligent data-processing algorithms used in RPM devices result in frequent false-positive alarms, resulting in alarm fatigue. This study aims to do a critical review of the existing literature to identify the causes of these false-positive alarms and categories the various interventions used in literature to eliminate these causes. That act as a catalog and help in false alarm reduction algorithm design. A step-by-step approach to building an effective alarm signal generator for clinical use has been proposed in this work. Second, the possible causes of false-positive alarms amongst RPM applications were analysed from the literature. Third, a critical review has been done for the various interventions used in the literature depending on causes and classification based on four major approaches: clinical knowledge, physiological data, medical sensor devices, and clinical environments. An effective clinical alarm strategy could be developed by following our pentagon approach. The first phase of this approach emphasises identifying the various causes for the high number of false-positive alarms. Future research will focus on developing false alarm reduction method using data mining.

# 1 Introduction

The internet of things (IoT) has improved the quality of medical care processes using innovative technology [1]. For progressive nations, human health and well-being are imperative goals. Medical devices used by health professionals and doctors are essential to monitor and prevent a patient's illness. IoT for patient monitoring provides several opportunities for researchers and developers to monitor patients remotely by using cloud and fog computing technologies [2,3]. The main characteristics of the IoT are [4]:

- Interconnectivity: IoT provides interconnectivity of most devices for communication and information exchange.

- Things-related services: The IoT can provide confidentiality and semantic consistency between physical and virtual things.

- Heterogeneity: Devices used in IoT are heterogeneous as each integrated device in this network is manufactured by a different vendor. Deployed intelligent devices can also transfer and gather data over the internet to reach their shared goal of providing the specialized service for which they are designed.

- Dynamic changes: The number of devices connected over the network, as well as their state of waking, active, sleeping, connected, and disconnected, is efficiently handled by the IoT.

- Enormous scale: The IoT handles data generated by many devices and their interpretation to generate useful information.

- Safety: This is one of the essential characteristics. The protection of our data, as well as the network endpoints, means creating a security paradigm.

- Connectivity: The connectivity network accessibility and compatibility of various things can be provided.

The IoT architecture consists of different technologies that safely deploy devices to deliver intelligent services. The sensors and actuators used in electronic devices are smart for communication and collection of data via network components. The collected data is transmitted to the storage servers for processing and then delivered as a service to the application users. Figure 1 shows how IoT communicates with other devices. All the recorded data is digital, which can be accessible by the end-users. The IoT architecture consists of the following four components:

- Smart Devices: A smart device is an electronic gadget that can connect, share, and interact with users and other smart devices. Although usually small in size, intelligent devices typically have the computing power of a few gigabytes.
- First Hope Network: The local network translates the device communication protocols to internet protocols.
- The Internet: The internet connects intelligent devices with backend servers.
- Back End Server: These are primarily data centers or client applications.

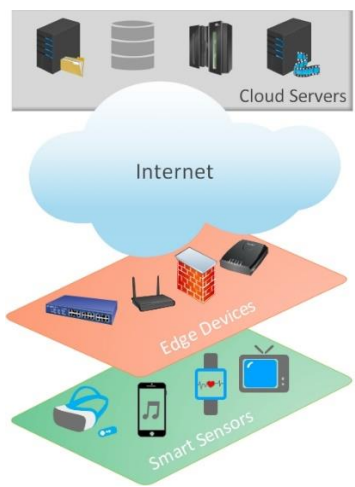

Figure 1. IoT Architecture

With the advancement of technology, IoT is the opening opportunity for many innovative applications such as healthcare, transportation, robotics, automation, logistics, industrial manufacturing. Remote patient monitoring (RPM) is one of the services provided by IoT that offers various benefits to both patients and physicians at anytime, anywhere, including continuous patient health monitoring and access to real-time health data through the internet [5,6].

## 1.1 Remote Patient Monitoring Application Architecture

Intelligent devices connected through networked systems can provide clinicians or caretakers with the patient's present and past health records in remote patient monitoring (RPM) applications. The RPM application architecture mainly monitors the patient's health condition and a processing part to identify health-related issues. The typical vital signs such as heart rate, blood pressure, temperature, respiratory rate, and $SpO_2$ can be measured by Bluetooth-enabled sensors that collect a patient's health data, as shown in Figure 2. The sensed data are pre-processed and sent to the connected gateway, usually a smartphone or tablet, via Bluetooth, which runs an application that authenticates the patient's id or other necessary information before transmitting to the cloud servers. After necessary authentication, the patients' real-time health data can be accessed using the cloud services from doctors or clinicians' personal computer or smartphones. The cloud services can have intelligent algorithms that can automatically raise alarms to clinicians for early intervention.

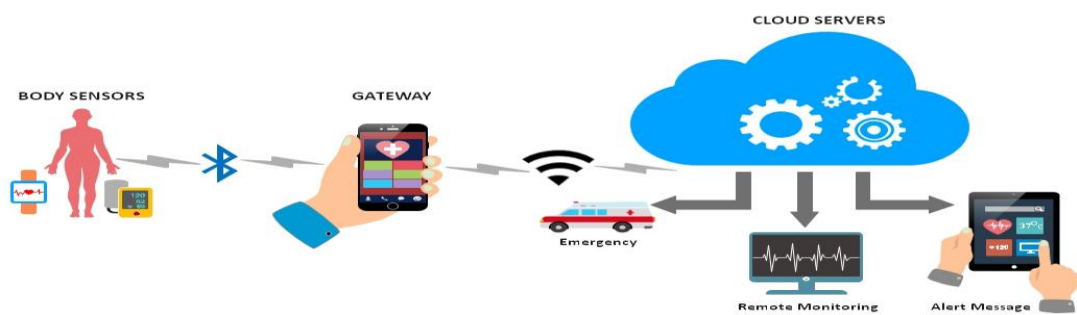

Figure 2. An Architecture for remote patient monitoring (RPM)

**1.2 Alarms in RPM**

The effectiveness of remote patient monitoring systems depends on algorithms for automatically processing data to raise clinically significant alarms for the clinicians to make an informed decision and early intervention [7]. These clinical alarm needs to be triggered 'right on time' and accurate so that treatment can be commenced 'well on time.' Automated alarm signals play a pivotal role in increasing the quality of care and contribute to patients' good health, especially in remote patient monitoring, where clinicians or guardians are away from the patient.

Monitoring devices raise alarms with advanced data processing algorithms for specific life-threatening situations when the life support devices are not working appropriately or when a patient's condition is forecast to deteriorate [8]. Therefore, an in-depth investigation of the patient's clinical needs and monitoring environment for efficient alarm generation is needed [7]. The capability of monitoring devices that detect a patient's clinical deterioration, depending on the algorithm used, results in frequent alarms. In most cases, most of these alarms may have crossed the pre-set parameter limits but have minimal or no clinical significance [9]. Due to frequent false positive alarms, clinicians can become less sensitive toward patients' alarms and neglect any possible dangerous situations [8], leading to alarm fatigue [8, 10, 11].

**1.3 False-Positive Alarms**

The literature in [12 -14] has identified that most of the alarms in Intensive care units (ICU) are non-actionable or false-positive, and only 5% – 13% of ICU alarms were actionable. One-quarter of monitored patients are responsible for two-thirds of all alarms [12]. In contrast, another study in [7] found that 2% of patients observed data in ICU contributed to 77% of false arrhythmia alarms. These studies show that only a few patients generate most false alarms.

Fewer false-positive alarms can be achieved in the ICU by applying the interventions to available medical devices such as changing the electrodes every day, with possible alarm delay, and the personalised widening of the alarm parameters as only < 26% of alarms found to be clinically significant [15]. Proper alarm settings such as turning off duplicate alarms and considering alarm delay for alarm autocorrection and staff education on setting personalised parameter limits are essential for reducing false alarms [16].

One of the significant issues with the current RPM application is the high sensitivity due to the advanced algorithm used in the medical devices, causing an increased number of false-positive alarms resulting in alarm fatigue, which decreases the quality of care [7, 17]. And these current RPM applications, historically, do not have the sensitivity and specificity of the commercial-grade, non-mobile sensors that the hospital used. This problem will likely be exacerbated as we migrate to more aggregated sensors with arguably worse patient/sensor interfaces such as skin contact. Various other factors such as the lack of standard guidelines, pre-set threshold, patient's movements, a faulty device, low battery, data-processing algorithms contribute to an increased number of unnecessary alarm signals [18-20].

The contribution of this chapter is that it considers all the literature that builds upon medical devices and allows the researcher to translate the knowledge for futuristic RPM applications. The lack of standard guidelines for generating the alarm signal has been identified in the literature. Our proposed pentagon approach to developing a more efficient alarm signal will attempt to address this gap in the literature. Various causes for the origin of false-positive alarms have been identified from the literature. A classification has been done for multiple interventions used in the literature to reduce false-positive alarms based on four major approaches: clinical knowledge, physiological data, medical sensor devices, and clinical environments focusing on the causes of false-positive alarms. This review was conducted according to the critical review guidelines [21].

The rest of the chapter is organized as follows: Section 2 describes the proposed pentagon approach. Section 3 presents the classification categories. Section 4 presents the results, and the discussion and conclusion are shown in the last section.

## 2  A Pentagon Approach

The literature does not address a general guideline to develop an effective alarm signal generation strategy for clinical needs. This work proposes the 'pentagon approach' for clinical alarm generation to guide researchers, clinicians, and developers to design an effective alarm generation strategy for their needs. An effective alarm generation strategy can help in reducing frequent false positive alarm signals. The proposed guidelines in the pentagon approach (see Figure 3) provide a well-structured flow of phases that help build an elegant solution at each level with clinical data verification. The guideline consists of the following five steps: the gathering of clinical requirements, selection of an appropriate method, application design and development, clinical trial, clinical analysis, and clinical response.

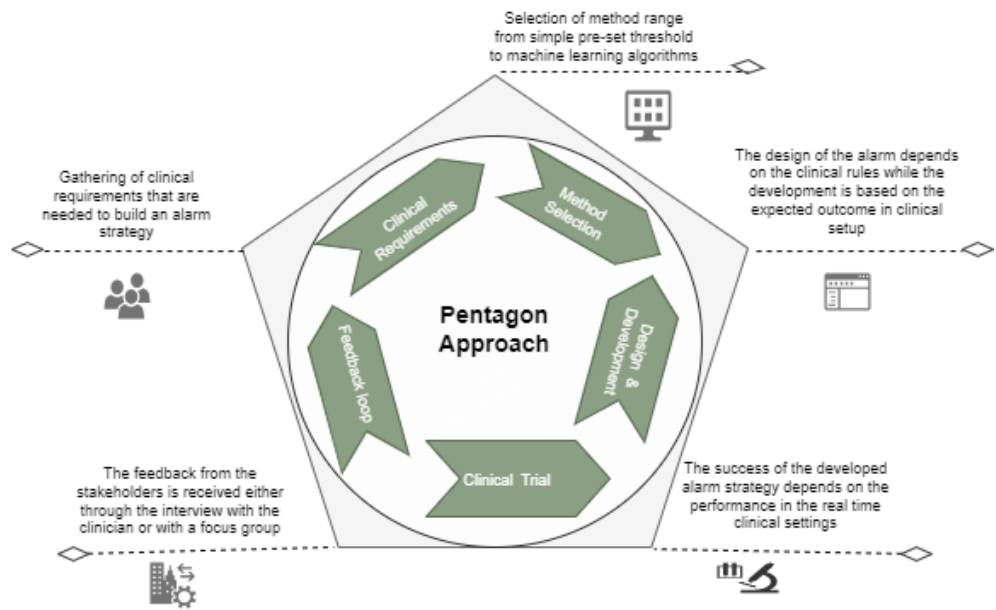

Figure 3. A pentagon approach to building a clinical alarm strategy

## 2.1 Clinical Requirements

This initial step in the guidelines includes gathering all the details required for the alarm strategy. A detailed list of clinical requirements should be made by interviewing the clinical staff and analysing the existing system. A few items on the clinical requirement list could be the scope for changes, the need for alarm, the expectation of patients and nurses, parameters to be observed, patient-centric alarm systems or generalised alarm systems, and the feasibility of the new systems, a scoring system to be used, medical devices to be used, architecture. Identifying the various causes that can affect the clinically recorded data value of the measured parameters is also essential for this phase. Studies show that some false-positive alarm signals are due to patient movement, smaller-sized sensors, or not appropriately placed sensors [22].

Selecting a suitable alarm strategy based on the gathered clinical requirements can help build an efficient system with fewer false-positive alarms. Literature [23] shows how identifying the causes and effects of unnecessary alarm signals is helpful for overall system quality. This paper proposed method selection focusing on the leading causes of the false-positive alarm in the current system. Identification of the leading causes for the origin of false-positive alarms has been discussed in section 3 in detail. Therefore, selecting methods is the following step after gathering the requirements in building an effective alarm signal strategy.

## 2.2. Method Selection

After gathering the clinical requirements and specifications, selecting the appropriate method to raise the alarm signal is imperative. The clinical needs and specifications list will input the proper alarm strategy phase. Method selection should be made to reduce the unnecessary alarms focusing on the clinical requirements. However, the literature does not identify various methods focusing on the alarm causes. A critical review has been done to identify and categorise the multiple methods used in the literature, focusing on the causes in this study to fill the gap. The critical review has been discussed in section 3. This categorisation will help researchers further identify the possible methods for the categories' causes.

The simplest and most widely used methods are customisation of the pre-set threshold value [24-27], machine learning techniques [28, 29], pattern matcher method [30], and median filter [31]. The time delay method of the sensor value received at the destination by the clinician or a caretaker in RPM using the time-critical parameter is a proper technique to lessen the anomalies in the alarm signal [32, 33].

### 2.3. Design and Development

Once the method is selected in the previous phase, the next step is to design and develop the alarm strategy. The design behind an alarm signal strategy depends on the clinical rules, algorithms, and necessary features. The development of the alarm signal strategy is carried out based on the expected outcomes, the risk associated with the alarm signals, security, and privacy of the clinical environment. In the case of remote patient monitoring, the design and development of the alarm signal, the strategy included networking technologies such as ZigBee, Bluetooth, and cellular networks [34, 35].

### 2.4. Clinical Trial and Analysis

The design and development from the previous phase for a required clinical setup take the form of software, smart device, or other tangible means. The developed solution is then implemented technically in the relevant clinical environments for the trial. Clinical trials are done to check the feasibility of the developed alarm signal strategy according to precise clinical requirements. The following aspects need to be considered for the clinical trial and analysis:

- various problems encountered
- alternate methods
- quality of applied methods
- reliability of the alarm
- validation of the alarming outcome
- performance measures

- limitation of the designed system in real-time

- software or hardware bugs

- improvements needed for various clinical needs

The literature shows that the developed alarm signal strategy is more effective if tested in the clinical setting before actual deployment [34, 36]. The reasoning alarm system designed for remote patient monitoring is more productive by testing the alarm system on various mobile devices and observing their response time [34]. The prototype of the developed fuzzy logic-based alarm system for automated anesthesia level controller has proven beneficial in literature [36].

### 2.5. Feedback Loop

After the clinical trial, the deployment of the alarm signal strategy is performed in relevant clinical settings. The feedback for these life-critical alarm systems is most effective through direct interviews with the clinicians, a Questionnaire, or a focus group of the stakeholders [37, 38]. Quantitative and qualitative approaches have a firm agreement and can be used as an effective tool for collecting feedback for these alarm systems. Key factors to be considered for this phase are as follows:

- Culture

- Ease of use

- Clinical staff perspective

- Patient perspective

- Security of data

- Performance

- Quality of results

Based on the collected feedback, the necessary steps are planned to make the system work better in the clinical environment. The motivation behind implementing the feedback is to maintain the implemented strategy for various new requirements using the pentagon approach. The following section describes the classification categories of multiple causes of false alarms.

## 3  Classification Categories

The first phase of the pentagon model is to identify the clinical requirements. Identifying the causes of false-positive alarms is one of the essential clinical requirements. The unusual movement of the patient [39], the vibration of the sensors, use of incompetent clinical knowledge, inappropriate medication, erroneous sensor placement on the body of the patient, time of the measurement, use of incorrect parameter values, incorrect threshold settings [40], false interpretation of the alarms, software bugs and hardware malfunction [41] are the few identified causes from the literature.

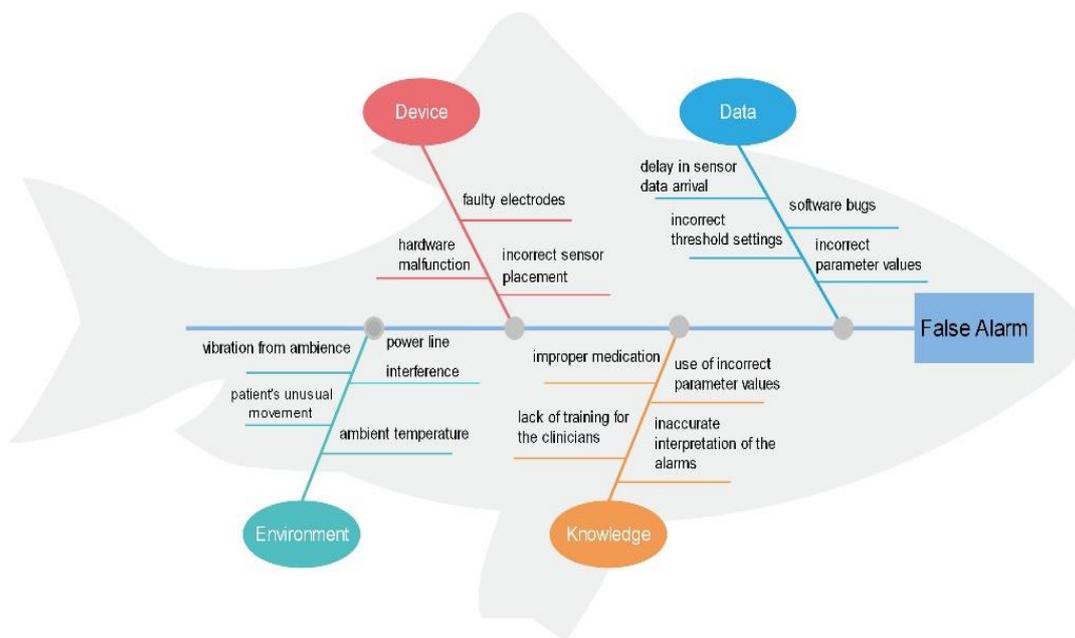

Figure 4. False alarm[1] causes in a clinical setup

The identified causes are categorised based on the nature of the cause of the false positive alarm under knowledge, data, device, and environment, as shown in Figure 4. As one of the main contributions of this work, we classified the causes into four broader categories. Our study of various alarm strategy methods in the literature convincingly falls under any one of these more general categories. Also, we have found that the causes of the alarm are related to one of these categories. The following gives a brief description of the categories and the nature of the causes of false-positive alarms.

---

[1] False alarm and false-positive alarm terms used in the paper signify the same meaning.

- *Data:* The false-positive alarms that arise due to incorrect parameter values, delay in data arrival of sensor data - in case of remote patient monitoring, incorrect threshold settings, and software bugs that directly affect the interpretation of the vital signs of the patient falls under the data category.
- *Device:* Device or technically related false positive alarms include faulty electrodes, erroneous sensor placement on the patient's body, and hardware malfunction that falls under the device category.
- *Environment:* False positive alarms can be triggered due to the unusual movement of the patient, external noise, power line interference, ambient temperature, and vibration from the ambience that originate in the clinical setup come under the environment.
- *Knowledge*: The knowledge category consists of those false positive alarms caused due to human error such as lack of training for the clinicians, inaccurate interpretation of the alarm signals, incorrect parameter values, and drastic changes in vital signs due to improper medication.

The motive behind categorising the causes of false-positive alarms is to identify the methods that can overcome the causes. The methods that are related to the causes are collated under approaches. The naming of the approaches is derived from the identified categories, as shown in Figure 5, and are based on the suitable method one can choose to reduce the effect of causes responsible for frequent false positive alarms.

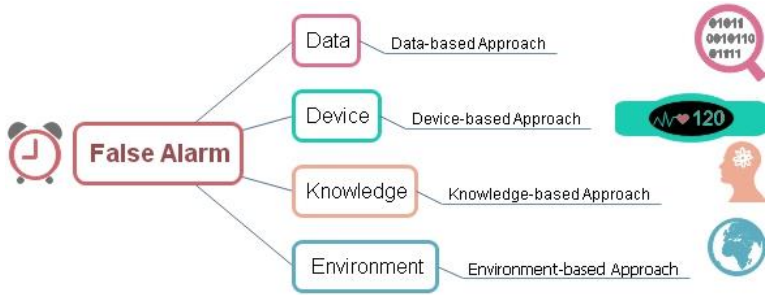

Figure 5. Classification of false-positive alarm[2] artefacts and approaches

Our literature review is based on the significant publications in the clinical and biotechnology field and further refined for the clinical trials. The literature review has shown the following methods used to reduce the number of false-positive alarms: pattern match, changing electrode, machine learning, time delay, median filter, dimension reduction, sensor fusion, integration, coalition game theory, and team-based. Figure 5 shows the approaches consist of different types of methods that are similar and related to the approach and are related to the approach type that can be used to solve the causes of the false positive alarm. The classification of the approaches and the methods is helpful in the second phase of the pentagon model, which is method selection, and will facilitate the third phase of the pentagon model is the design and development of the alarm signal strategies depending on the nature of the false alarm. A detailed description of the approach and the methods are given in the following section.

### 3.1. Physiological Data-based Approach

In physiological data-based approaches, the observed vital signs data is the leading cause of false alarms. The identified methods used to reduce clinical false alarms focusing on the physiological data are discussed below.

---

[2] False alarm and false-positive alarm terms used in the paper signify the same meaning.

*3.1.1. Customised Alarm signals*

The customised alarm signals are those alarm signals that are set manually based on user requirements and differ based on the user's physiology and medical condition. Studies show that personalised alarm signal parameters and set manual limits for physiological monitoring reduced false-positive alarms by 43% from the baseline [24]. For instance, 89% reduction in the total mean weekly alarm signals were measured by self-reset alarm signals for bradycardia, tachycardia, and heart rate (H.R.) limits where low H.R. is et to 45 bpm and high H.R. is set to 130 bpm [25]. Adapting parameter limits of $SpO_2$ monitors also results in fewer warnings per day and improved' nurses' satisfaction with the less frequent and effective alarm signals [26]. Notable improvements were achieved by implementing procedure-specific vital signs settings for cardiac patients by considerably reducing 80% of false-positive alarms [27].

*3.1.2. Machine Learning*

In machine learning methods, alarm signal systems learn about a clinical event that needs to be detected with the help of customised algorithms and datasets. Multivariate approaches in ICU alarm signals can achieve higher accuracy than using different algorithms for different alarm signals. These kind-of approaches can extract hundreds of relevant features to capture the characteristics of all alarm signals from arterial blood pressure (ABP) and electrocardiogram (ECG). The comparison study [28] found a 13.96% false-positive alarm suppression rate against feature-based false-positive alarm detection.

Srivastava [29] shows that feature extraction based on data segments has been used for machine learning. Based on this method, two models using random forest classifiers (RFC) and thresholds have been developed to assess the data. K-fold cross-validation has been applied for the evolution of RFC due to less than 750 records from the physio-net database. Simple association rules have been used on two designed models using the two subsets of data for arrhythmia and predicted 95.56% of reliable alarm signals.

The machine-learning algorithm was applied to 4107 'experts' labeled data extracted from 114 quality signals of ECG, PPG, and ABP waveforms. The false-positive alarm suppression rate for the extracted dataset is 86.4% for asystole, 100% for extreme bradycardia, and 27.8% for severe tachycardia with no suppression of any true-positive alarms. The suppression of true-positive alarm means that reliable alarm signal or clinical usefulness was not suppressed. While for the ventricular tachycardia alarm signals, the false-positive alarm suppression performance was 30.5%, with a true-positive alarm suppression rate below 1% [42].

*3.1.3. Time delay*

The time delay method uses some delay time before triggering the alarm signal. This delay proved helpful in false-positive alarm reduction. Decreasing the alarm limit of SpO2 to reduce false-positive alarms may increase more episodes of hypoxemia. However, a delay of 15 seconds between crossing the threshold value and triggering the alarm signal would reduce the false-positive alarm by 60% [15]. A minimal threshold variation of short durations of a 14-second delay in an ICU reduced false-positive alarms by 50%, while a 19-second delay reduced alarms by 67% [43].

Over 12 months, 316,688 alarms were recorded for 6701 patients in trauma resuscitation. It was found that a 2-second delay in alarm signal would reduce incidents by 25%, and 5 seconds would reduce the alarm signals by 49% [44].

*3.1.4. Integration of techniques*

The integration of techniques uses sensor fusion, time delay, feature extraction, etc., to reduce false-positive alarms. For example, the algorithm derived from the features and timings of the ABP signals used to reduce ECG arrhythmia alarm signals on an average of 47% for 447 adult patient records in the MIMIC II database [45].

Suppressing ICU alarm signals using automated features and engineering using ABP signals, coupled with machine learning algorithms such as vector machines, random forest, and extreme random tree classifier, suppressed 90.3% of the false tachycardia alarm signals. And only 0.54% of the true-positive alarms were incorrectly suppressed [17].

*3.1.5. Coalition game theory*

Coalition game theory works by grouping the agents in the game together. A similar way for a patient monitoring system based on this theory has been developed, which considers inter-feature dependencies on the PhysioNet's MIMIC-II database also resulted in an improved classification of the alarm signals [46]. Inter-feature's mutual information helps in the accuracy of the classification of the alarm signals. This method has been applied to collect data from hospitals for arrhythmias, including asystole, extreme bradycardia, extreme tachycardia, ventricular tachycardia, ventricular flutter/fibrillation, arterial blood pressure (ABP), and Photoplethysmogram (PPG). And observed features such as mean, variance, median, kurtosis, and entropy resulted in the successful suppression of 75% of false-positive alarms [47].

*3.1.6. Sensor fusion*

Different physiological sensor information is fused in sensor fusion. This information was then fed into the customised algorithm to predict true-positive alarms. A set of algorithms using heart rate variability (HRV) index, Bayesian inference, neural networks, fuzzy logic, and majority voting has been proposed in [48] to fuse the ECG, arterial blood pressure, and PPG. Upon fusion, three kinds of information are extracted from each source. Namely, heart rate variability, the heart rate and difference between sensors, and the spectral analysis of low and high noise of each sensor are fed as an input to the algorithms. The results are validated with twenty sets of recordings from the MIMIC database, which showed that neural networks fusion had the best false-positive alarm reduction of 92.5 %. In contrast, the Bayesian technique decreased by 84.3 %, fuzzy logic was 80.6 %, majority voter was 72.5 %, and the heart rate variability index showed a 67.5 % reduction [48].

A Bayesian approach has been used to fuse electrocardiogram, arterial blood pressure, and PPG from physiological sensors to create robust heart estimations. The HRV index and majority voter technique are then compared using 20 selected records from the MIMIC II database. Results showed that Bayesian fusion presents a lower error rate of 23%. In comparison, the other evaluated techniques give an error rate of 35% (ECG Only), 40% (ABP Only), 41% (PPG Only), 37% (HRV Index), and 31% (Majority Voter) [49].

*3.1.7. Pattern discovery*

Pattern discovery identifies various signal patterns by observing physiological data features, time, the relationship between alarm signal categories, etc., which are responsible for alarm signals. Heuristic techniques to extract inter alarm signal relationships include identifying the presence of alarm signal cluster, the pattern of transition from one alarm signal category to another, temporal association, and prevalent sequence have developed. Desaturation, bradycardia, and apnea constitute 86% of alarm signals, and by inhibiting further of a category, the 30s/60s alarm signal reduced by 20% [50].

## 3.2. Clinical Device-centric Approach

Artefacts[3] caused due to device-related issues falling under clinical device-centric approaches. The approaches for the solution should be device-based for causes such as faulty devices, device malfunction, defective device parts, and aged devices. Literature shows that few studies have focused on devices to reduce false-positive alarms. Changing the electrodes daily on cardiac monitors results in average alarm signals per bed being reduced by 46% [51]. A two-adult nursing unit was examined by changing the electrocardiographic electrode for two intervention periods – the change of electrodes showed the cardiac monitor alarm events were decreased by 20.6% and 71.0% in the first and second intervention periods [52].

---

[3] Artefacts are anomalies introduced into digital signals because of digital signal processing.

### 3.3. Clinical Knowledge-based Approach

An erroneous assessment of the patient's condition or incorrect interpretation of the false-positive alarm falls under clinical knowledge-based approaches. The approaches that use the knowledge gained from experience or research to interpret the alarm signal fall under this category.

*3.3.1. Pattern match*

The clinician uses pattern matching to match patterns from the observed data or signals. A knowledge-based approach, based on pattern match, for neonatal intensive care unit for probe change has been conducted in [53] and shows that 45 probe changes with the accuracy to identify 89% of false-positive alarms. The clinician marks up features of interest on the monitor data. Then their knowledge is used as a pattern matcher over a particular set of intervals for a new collection of raw data [54].

*3.3.2. Team-based method*

Grouping various methods based on clinician knowledge can be done, which acts as a team to reduce false-positive alarms. A team-based method to reduce cardiac monitor alarms has been developed. This method has four steps:

- Setting up age-appropriate parameters
- Daily electrode changes
- Individual monitor assessment
- Procedure to stop monitoring

The results show compliance increased from 38% to 95%, and the median number of alarms decreased from 180 to 40, which is a 77.78% decrease in false alarms [55].

### 3.4. Clinical Environment-based Approaches

In this approach, artifacts are due to the surrounding environment in which the patient monitoring setup resides. The causes such as power line interference, electrode malfunction, floor vibrations, wobbly connections, and any other artefacts caused due to multiple sources that may or may not be related to other sources fall under this category.

*3.4.1. Median filters*

A median filter removes short-term noise in the measurement signal without manipulating the baseline signal. Some literature studies use this method to remove interference caused by sources like power lines, noise, subject movement, etc. Due to power sources in the ECG signal, interference could be reduced using the notch filter in the monitors, a band-stop filter with a narrow stopband is used [56].

An effective combination of a "short" (15 seconds) and a "long" (2.5 minutes) filter in a database of ten cardiac surgery patients has been evaluated. The result showed that therapeutic consequences increased from 12 to 49 %, with no relevant alarm signal missed [31].

The artificial neural network (ANN) based algorithm has been developed for removing non-linear time-varying noise characteristics of ECG to detect QRS complex. The noise removal using a linear whitening filter for MIT/BIH arrhythmia database records 97.5%, and band-pass filtering records 96.5% [57].

*3.4.2. Dimension reduction*

Dimension reduction reduces the number of input variables used to evaluate the output, as more inputs make the processing task more complicated. Continuous and non-invasive B.P. monitoring problems have been addressed with a proposed methodology of de-noising, feature extraction, and various regression stages using dimension reduction. For dimension reduction, the study reduced the length of the original feature vectors using Principal Component Analysis (PCA) to reduce the feature-length from over 190 to 15. Using this methodology, the cumulative error percentage for threshold 15mmHg comes from diastolic blood pressure (DBP) 95.7%, mean arterial pressure (MAP) 93.1%, and systolic blood pressure (SBP) 72.7%, which is less as a comparison to British hypertension society standards [58].

## 4 Results and Discussion

We have reviewed the various methods and techniques to reduce false-positive alarms in patient monitoring systems. The leading causes of false-positive alarms in these methods and techniques lack analysis of the sources of false-positive alarms. There are various artefacts like movement, vibration, and noise in the clinical setup - these artifacts are the reason for false-positive clinical alarms [59]. The recorded artefacts can lead to changes in physiological recorded patient data or misinterpret the algorithm used for monitoring patient's conditions, resulting in higher numbers of false-positive alarms.

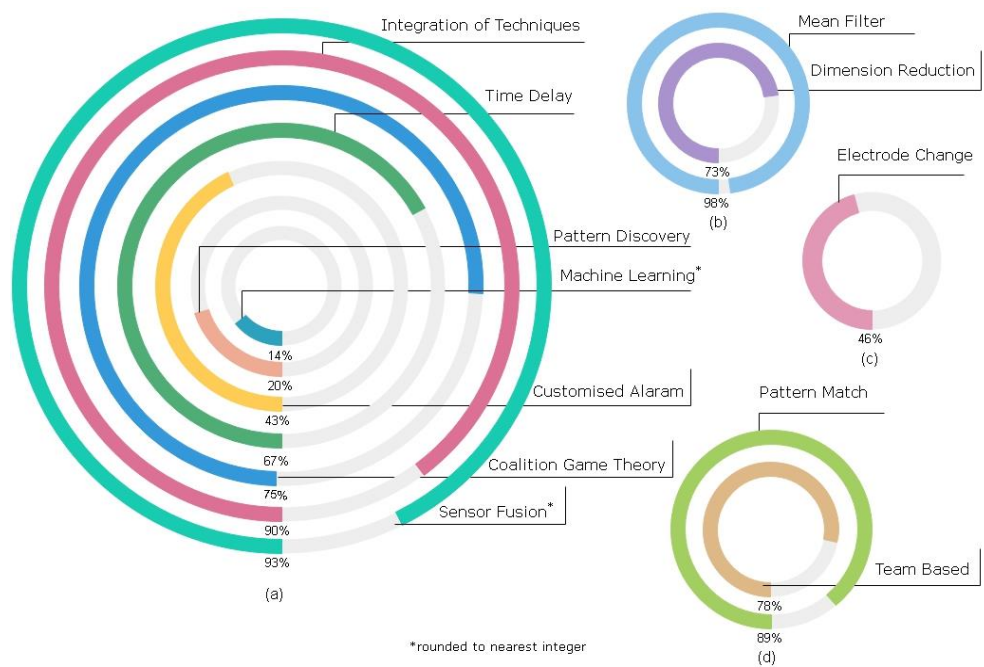

Figure 6: Suppressed false-positive alarm rate for (a) Physiological Data (b) Clinical Environment, (c) Clinical Device Centric, and (d) Clinical Knowledge-based approaches.

We have categorised various approaches based on the nature of the causes that have been identified in the literature to reduce false-positive alarms. The graph in Figure 6 has been plotted for suppressing false-positive alarm rates in the proposed approaches by various methods and techniques in these approaches. It is not feasible to collate all the methods that use the same sample size and clinical setup. Therefore, this paper considers all those methods that conducted clinical trials rather than simulations.

### 4.1. Physiological Data

The chart shows that most false-positive alarms are due to the vital signs data. Subsequently, many methods suppress the same, such as customised alarm parameters, machine learning, time delay, coalition game theory, integrating various methods, sensor fusion, and pattern discovery. These listed methods focus on patient data to reduce false-positive alarms using multiple ways.

### 4.2. Clinical Device-Centric

Interestingly, the number of false-positive alarms due to the device is minimal. Subsequently, daily electrode change is the only method found in the literature to suppress the false-positive alarm caused due to device-centric issues. The apparatus used for remote patient monitoring is getting more reliable and efficient. Changing the electrode daily effectively suppressed false-positive alarms for device-related problems by 46%.

### 4.3. Clinical Knowledge-based

Similarly, fewer clinical false-positive alarms are found due to incomplete clinical knowledge. The pattern match is one of the methods under clinical false-positive alarms with the suppress false-positive alarm rate of 89%. Clinicians use their knowledge to identify the patterns in the recorded patient data to predict the patient's condition. Another method under clinical knowledge is the team-based method, with suppresses a false-positive alarm rate of 78%.

Clinicians use their knowledge to use various methods such as customised parameters or electrode changes to suppress the false-positive alarm. This clarifies minimum human error in the healthcare environment and the importance of regulatory measures followed in the device certification.

### 4.4. Clinical Environment

Although the clinical setup also contributes towards the false-positive alarms and the related methods to reduce them are median filters, which filter the signal for noises or disturbances, and the false-positive alarm suppresses rate found to be 96%. Dimension reduction methods were found to suppress the false-positive alarm rate by 73%.

We have found that most artefacts causing false-positive alarms affect clinical recorded data with the overall observation. Many effective methods have been listed under a clinical data-based approach that focuses on clinical data to suppress false-positive alarms. Under clinical data-based approaches, sensor fusion and integration of various techniques have been found effective as they could suppress false-positive alarm rates of more than 90%. The clinical devices became more efficient and reliable using updated hardware and software, causing fewer false-positive alarms. On the other hand, when the clinician uses their knowledge to match the pattern to reduce the false-positive alarm, they suppress the false-positive alarm rate by 89%. As fewer clinical alarms are recorded in the clinical setup, the mean filter is one of the effective methods to suppress false-positive alarms.

## 5. Conclusions

The presented pentagon clinical alarm generation approach gives the guidelines with five phases that will be highly useful for the applied researchers and developers to create an efficient clinical alarm signal strategy for patient monitoring. Significant factors to suppress false-positive alarms in the alarm strategy are highlighted in the pentagon approach. A critical review of various interventions from literature for false-positive alarms has been done to categorise them based on physiological data used, clinical device involvement, use of clinical knowledge, and the setting up of the clinical environment. These approaches are classified based on the source for various artefacts responsible for the false-positive alarms. This classification will act as a catalog for the researchers to view multiple interventions used in literature focusing on the artefacts for the false alarms. Our proposed work articulated that the methods used to reduce false clinical alarms without analysis of artefacts are significant sources of frequent false-positive alarms. For our future research direction, better physiological data-based approaches with the help of data-mining techniques will be designed to reduce the number of false alarms in an RPM.